\documentclass[letter, 10 pt, conference]{ieeeconf}
\IEEEoverridecommandlockouts
\usepackage{comment}

\usepackage{graphics} 

\usepackage{amsmath}
\usepackage{caption}
\usepackage{fancyref}
\usepackage{subcaption}
\usepackage{multirow}
\usepackage{graphicx}
\usepackage{xcolor}
\usepackage[ruled,vlined,linesnumbered]{algorithm2e}
\SetKwInput{KwInput}{Input}
\SetKwInput{KwOutput}{Output}
\SetKwInOut{KwParam}{Parameter}
\usepackage{hyperref}
\usepackage{tabularx}
\usepackage{cite}
\DeclareMathAlphabet{\mathcal}{OMS}{cmsy}{m}{n}
\usepackage{epsfig} 
\usepackage{mathptmx} 
\usepackage{times} 
\usepackage{amsmath} 
\usepackage{amssymb}  
\usepackage{pifont}
\newcommand{\zono}[1]{\left\langle#1\right\rangle}

\newtheorem{definition}{Definition}
\newtheorem{proposition}{Proposition}

\newcommand{\R}{{\mathbb{R}}}

\usepackage[compact]{titlesec}
\titlespacing{\section}{0pt}{1.3ex}{1.3ex}
\titlespacing{\subsection}{0pt}{0.6ex}{0.6ex}
\titlespacing{\subsubsection}{0pt}{0.4ex}{0.4ex}

\title{\LARGE \bf Shared Situational Awareness with V2X Communication \\
and Set-membership Estimation}
\author{Vandana Narri$^{1,2}$, Amr Alanwar$^{3}$, Jonas Mårtensson$^{2}$, Christoffer Norén$^{1}$ and Karl Henrik Johansson$^{2}   $
\thanks{This paper has been submitted to IEEE Intelligent Transportation Systems 2023 and is currently under review.}%
\thanks{$^{1}$The authors are with Research and Development, Scania CV AB, 151 87 Södertälje, Sweden. {\tt\small \{vandana.narri, christoffer.noren\}@scania.com.}}%
\thanks{$^{2}$The authors are with Division of Decision and Control Systems at School of Electrical Engineering and Computer Science, KTH Royal Institute of Technology, SE100 44 Stockholm, Sweden. {\tt\small\{narri, jonas1, kallej\}@kth.se.}}%
\thanks{$^{3}$The author is with the School of Computer Science and Engineering, Constructor University, Germany {\tt\small aalanwar@constructor.university.}}
}

\begin{document}

\maketitle
\begin{abstract}

The ability to perceive and comprehend a traffic situation and to estimate the state of the vehicles and road-users in the surrounding of the ego-vehicle is known as situational awareness. Situational awareness for a heavy-duty autonomous vehicle is a critical part of the automation platform and depends on the ego-vehicle's field-of-view. But when it comes to the urban scenario, the field-of-view of the ego-vehicle is likely to be affected by occlusion and blind spots caused by infrastructure, moving vehicles, and parked vehicles. This paper proposes a framework to improve situational awareness using set-membership estimation and Vehicle-to-Everything (V2X) communication. This framework provides safety guarantees and can adapt to dynamically changing scenarios, and is integrated into an existing complex autonomous platform. A detailed description of the framework implementation and real-time results are illustrated in this paper.

\end{abstract}

\section{Introduction}
\label{sec:intro}

Many industries break the Connected Automated Vehicle (CAV) framework functionality into sensing, localization, perception, situational awareness, planning, and control. When CAV is located in an urban scenario, the field-of-view (FOV) of the CAV is severely affected by occlusions which leads to limited information about the surroundings and affects situational awareness of the CAV. To be able to overcome the challenge of limited information in an urban setting, the plan is to use Vehicle-to-Everything (V2X) communication to share perception data between road-users.

V2X communication is synonymous with Vehicle-to-Vehicle (V2V) and Vehicle-to-Infrastructure (V2I) communications. These communications use the ad-hoc wireless technology ITS-G5 (a.k.a. Dedicated Short-Range wireless Communication (DSRC) or IEEE~802.11p in the US). There are three different types of V2X communication today: ad-hoc (the focus of the present article), cloud-based, i.e., cloud-to-cloud communication, and cellular-assisted. The main strength of ad-hoc communication is the possibility to communicate everywhere and anytime without base station coverage, e.g., on rural roads and missing network coverage. In this paper, we propose a framework that can provide safety guarantees and increase situational awareness using set-membership estimation and ad-hoc V2X communication.



\begin{figure}[t]
    \centering
    \frame{\includegraphics[width=1\linewidth]{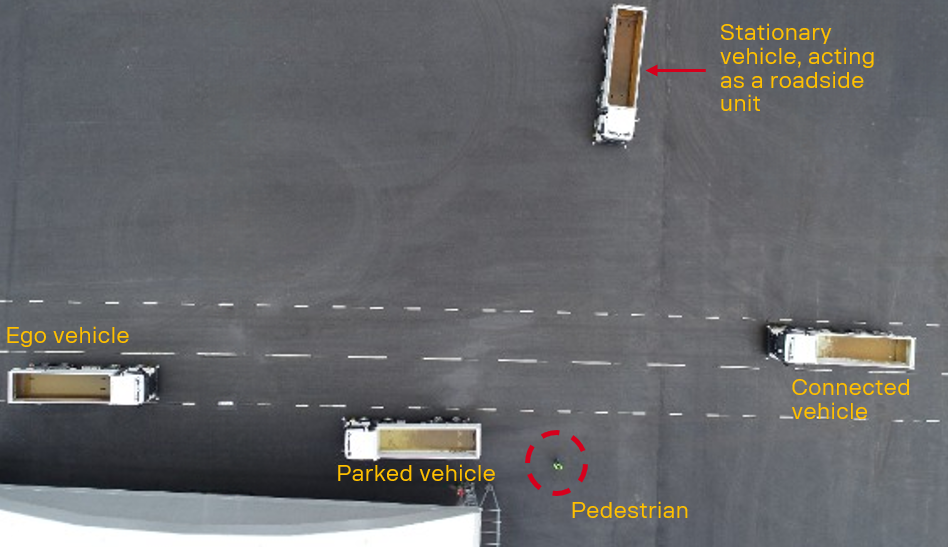}}
    \caption{This is a drone shot of the scenario with four vehicles and a pedestrian. The pedestrian is in the occluded area of the ego-vehicle. This picture is taken at the Scania test track.}
    \label{fig:scenario_realtime}
    \vspace{-6mm}
\end{figure}

\subsection{Literature Review}

Situational awareness of the Ego-Vehicle~(EV) depends on the perception module, which primarily relies on numerous onboard sensors, such as LiDAR, cameras, and millimeter-wave radar, but still lacks accuracy and complete information~\cite{wang2019multi}. One way to overcome this problem is by using V2X communication. In~\cite{cui2022cooperative}, the authors provided a detailed review of cooperative driving by introducing current cooperative perception information fusion methods and information-sharing strategies. In~\cite{yoon2021performance}, authors have provided a generalized cooperative perception framework based on decentralized V2V vehicular communication and presented evaluations of the framework based on randomized traffic simulations for multi-lane highway and roundabout intersection scenarios, considering both communication losses and sensor FOV resolution issues. Another related work in \cite{InformationEntropy} introduces information entropy, which quantifies uncertain information in the blind area, into the motion planning module of autonomous vehicles. Then, the authors propose to plan collision-free trajectories using model predictive control. 

In~\cite{mo2022method}, a framework of vehicle-infrastructure cooperative perception and data fusion method is proposed along with an improved Kalman filter when there is a roadside unit failure. In~\cite{queralta2019collaborative}, authors presented an architecture for coordinating an autonomous team of heterogeneous aerial and land robots that work together on collaborative mapping. Based on the presented concept, they proposed an IoE (Internet-of-Everything) architecture having heterogeneous support units for enhancing the situational awareness of autonomous vehicles in an unknown environment. A set-membership state estimator for autonomous surface vehicles along with dynamical decomposition that decouples the estimation problem for the rotational and positional dynamics is proposed in~\cite{CombastelSurfaceVehicles}. In~\cite{yee2018collaborative}, authors have proposed a novel data fusion algorithm by implementing the latest deep learning techniques and aggregating object (road-user) detection information from multiple viewpoints to improve detection performance. The objective of~\cite{9709015} is to provide a preliminary study of edge-assisted collaborative perception in autonomous driving from a V2X communication design perspective.



\subsection{Contributions}

Situational awareness is the ability to perceive and comprehend a traffic situation and estimate the state of road-users in the surrounding of the EV. The traditional way of the situational awareness module is to rely entirely on onboard sensor perception, also known as local perception. This work aims to use V2X communication to receive external perception, i.e., perception from other CAV or Road-Side Units (RSUs), to create shared situational awareness, which helps improve the understanding of one's surroundings and guarantee safety. Inclusion of perception obtained via V2X communication, helps in widening EV's FOV and creating line-of-sight in occluded areas. But external perception can induce varying unknown uncertainties. 

One of the main challenging parts is integrating the new functionality, i.e. shared situational awareness that relies on local and external perception into the existing CAV system architecture of the EV. This paper focuses on the fusion of local and external perceptions, which have varying uncertainties and involves different coordinate frames, to create a shared situational awareness for the EV using set-membership estimation. Set-membership estimation estimates a set that contains the state with set containment guarantees. In this paper, we have considered a scenario in which the pedestrian is occluded by the parked vehicle in the EV's FOV, as presented in Fig.~\ref{fig:scenario_realtime}. 
This is an extended work along with real implementation on Scania test track, from our paper that highlighted the concept of shared situational awareness using V2X communication~\cite{9575828}.

\subsection{Outline}
The paper is organized as follows: In Section~\ref{sec:pb}, the problem formulation is described. The necessary preliminaries are provided in Section~\ref{sec:preliminary}. The proposed framework is described in Section~\ref{sec:mehto}. Then, the evaluation section is provided in Section~\ref{sec:evaluation}, followed by Section~\ref{sec:conc} to conclude the paper. 


\section{Problem Formulation}
\label{sec:pb}

In this paper, the problem of occlusion and blind spots caused by infrastructure, moving vehicles, and parked vehicles to the EV is considered. A specific scenario is considered for the experiment, which is shown in Fig.~\ref{fig:scenario_realtime}. All the vehicles 
present in Fig.~\ref{fig:scenario_realtime} are equipped with a V2X communication module, which allows us to share perception information between all the road-users in the scenario except the parked vehicle and the pedestrian. 
It is also considered to have a distributed system, where the communication is set up as a peer to peer network and does not rely  upon communication infrastructure such as access points or base stations. The objective is to improve situational awareness of the EV. The following research questions are formulated to solve the problem in this paper:
\begin{enumerate}
    \item How to obtain and fuse data from local and external sensors to improve the situational awareness of the EV?
    \item How to provide safety guarantees of the fused data under varying measurement uncertainties?
\end{enumerate}

\section{Preliminary}
\label{sec:preliminary}
This section gives a detailed description of the system state and measurement models for each road-user. Then, we introduce zonotopes, which are used to represent the reachable set mathematically and followed by the system architecture of a CAV. 

\subsection{System State and Measurement Model}

This paper uses a discrete-time linear system to describe the state model for the road-user observed in the given scenario. The state vector consists of the position and velocity of the road-user as described in~\eqref{eq:sysmodel}. All the road-users state vectors are calculated with respect to the EV coordinate system. For $k = 1,2,\dots, T$, where $T$ is the time horizon, the state model can be described as follows.


\begin{equation}
\label{eq:sysmodel}  
x^j_{k+1} = F^j_k x^j_{k} + q^j_k,
\end{equation}
with
\begin{equation}
    \label{eq:systemstatevector}
    x^j_{k} = {\begin{bmatrix}
        x_{\text{x}} & x_{\text{y}} & x_{\text{s}}
    \end{bmatrix}^j_{k}}^\top,
\end{equation}
where $x^j_{k} \in \mathbb{R}^{n}$ with $n=3$; is the state vector of the $j^\text{th}$ road-user with $x^j_{0}$ as the initial state and $j = 1,\dots,n_r$, $n_r$ is the number of road-users, $x_{\text{x}}$, $x_{\text{y}}$, and $x_{\text{s}}$  are the \text{x} coordinate, \text{y} coordinate, and the velocity of the $j^{th}$ road-user respectively, $q^j_k$ is the process noise, and $F^j_k \in \mathbb{R}^{n\times n}$ is the state matrix at time $k$ which is defined as follows:
\begin{gather*}
     F^j_k = \begin{bmatrix}
        1 & 0 & \alpha_k^j  \\
        0 & 1 & \gamma_k^j  \\
        0 & 0 & 1   
    \end{bmatrix}
\end{gather*}
where $\alpha_k^j = \Delta t\cos{\theta_k^j}$, $ \gamma_k^j = \Delta t \sin{\theta_k^j}$. Here $\theta_k^j$ is the heading of the road-user at time $k$ and $\Delta t$ is the time step. 

The observable discrete-time linear system is assumed to generate the sensor measurements. The measurement model for the sensor is described as the following:
\begin{equation}
\label{eq:senmodel}
    y^j_{k} = H^j x^j_{k} + v^j_{k},
\end{equation}
with
%
\begin{equation}
 \label{eq:measurementstate}
 y^j_{k} = {\begin{bmatrix}
     y_\text{x} &  y_\text{y} &  y_\text{s}
 \end{bmatrix}^j_{k}}^\top,
\end{equation}
where $y^j_{k} \in \mathbb{R}^{p}$ with $p = 3$; is the measurement vector, which consists of  $y_\text{x}$, $y_\text{y}$ and $y_\text{s}$ i.e. $\text{x}$ coordinate, $\text{y}$ coordinate and velocity of the $j^\text{th}$ road-user at time step $k$ with respect to the EV coordinate system and $v^j_k$ the measurement noise. The measurement matrix is give by $H^j \in \mathbb{R}^{p \times n}$ and in this case it is an identity matrix. Since the processing of all the measurements will be done with respect to the local coordinate system of the EV, all the measurements obtained externally are converted to the local coordinate frame of the EV, which are denoted by $\text{x}_E$ and $\text{y}_E$, where $E$ subscript refers to the EV.


\subsection{Zonotopes}
There are various types of set representations, such as ellipsoids~\cite{bolting2019iterated}, polytopes~\cite{blesa2012robust}, zonotopes~\cite{conf:dis-diff}, orthotopes~\cite{belforte1990parameter} and intervals~\cite{xu2020interval}. In this paper, we use zonotopes to represent the set mathematically due to its special computational efficient properties. The definition and properties of zonotopes are stated next.

\begin{definition}[\textbf{Zonotope \cite{conf:zono1998}}] A zonotope $\mathcal{Z}= \zono{ c,G }$ consists of a center $c \in \R^n$ and a generator matrix $G$ $\in$ $\R^{n \times e}$. The generator matrix $G$ is composed of $e$ generators $g_{i} \in \R^n$, $i=1,..,e$, where $G=[g_{1},...,g_{e}]$ and $\beta_i$ is a scaling factor. Hence, 

\begin{equation}
     \mathcal{Z} = \Big\{ c + \sum\limits_{i=1}^{e} \beta_i  g_{i} \Big| -1\leq \beta_i \leq 1 \Big\}.
    \label{equ:zonoDef}
\end{equation}
\end{definition}


Zonotope can be interpreted as the Minkowski sum of a set of line segments. Given two zonotopes $\mathcal{Z}_1=\langle c_1,G_1 \rangle$ and $\mathcal{Z}_2=\langle c_2,G_2 \rangle$ and a scalar $L$, the following operations can be computed exactly \cite{conf:zono1998}:
\begin{itemize}
    \item Minkowski sum:
    \begin{equation}
     \mathcal{Z}_1 \oplus \mathcal{Z}_2 = \Big\langle c_1 + c_2, [G_1, G_2]\Big\rangle.
     \label{eq:minkowski}
     \end{equation}
    
    \item Scaling:
    \begin{equation}
     L \mathcal{Z}_1  = \Big\langle L c_1, L G_1\Big\rangle. 
     \label{eq:linmap}
     \end{equation}    
\end{itemize}

\subsection{CAV System Architecture}
\label{subsec:exist_arch}
\begin{figure}[t]
    \centering
    \includegraphics[width=\linewidth]{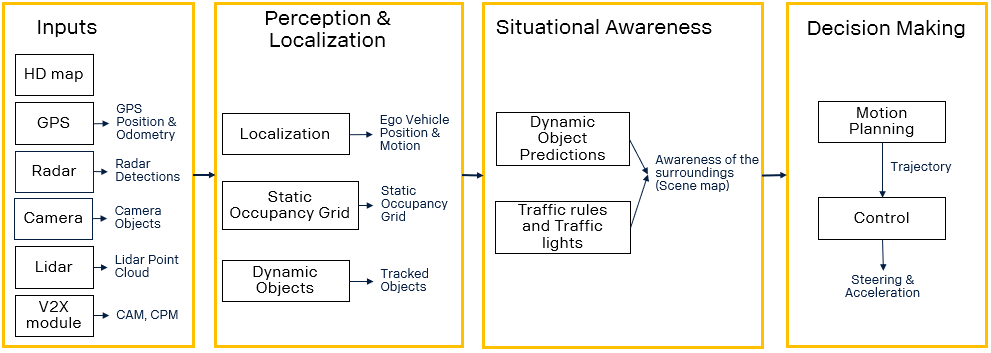}
    \caption{Existing system architecture of a CAV.}
    \label{fig:existing_arch}
    \vspace{-4mm}
\end{figure}

A CAV has a complex system architecture consisting of multiple sub-systems handling various tasks that enable the CAV to drive with less human interaction. The term connected states that the vehicle is capable of communicating with roadside units or other CAVs. An overview of an existing system architecture of a CAV is presented in Fig.~\ref{fig:existing_arch}. The system architecture is divided into four sub-systems: input layer, perception and localization layer, situational awareness layer, and decision-making layer.

The input layer consists High-Definition (HD) map, GPS, lidars, radars, cameras, and a V2X module that enables the communication between other agents. This input layer is essential and supports the functionality between different applications in the system. The perception module is responsible for detecting, identifying, classifying, and tracking different road-users in the environment surrounding the vehicle. This module is divided into two parts: static occupancy grid and dynamic objects. The static occupancy grid provides the FOV of the EV, and the dynamic objects provide the perception data, i.e., a list of objects detected in the FOV of the EV. The vehicle localization module is responsible for estimating the state of the vehicle (e.g., position, speed, orientation, and acceleration), both with respect to its surroundings (locally) and with respect to a map (globally). This module is essential for safe, comfortable, and precise vehicle motion.

The objective of the situation awareness module is to determine the predictions of the surrounding road-user's actions and provide scene understanding. The motion planning module's responsibility is to compute a 
trajectory while respecting vehicle dynamical constraints, being safe and smooth, and making the motion comfortable for passengers.
The motion control module's responsibility is to stabilize and guide the vehicle toward a given reference path or trajectory. The vehicle controller has to handle disturbances to the vehicle, correcting its states (e.g., position, yaw, and velocity) back to the desired reference while maintaining the actuation limits and achieving a smooth and comfortable ride.

\section{Methodology}\label{sec:mehto}

In this section, the architecture of the proposed framework is discussed, and each module of the framework is explained in detail. The proposed framework extends the conventional control architecture of an automated vehicle~\cite{gonzalez_review_2016} by using the set-membership method instead of statistical estimation for sensor fusion in order to have set containment guarantees which paves the way for safety guarantees.
\begin{figure}[b]
    \centering
    \includegraphics[width=0.9\linewidth]{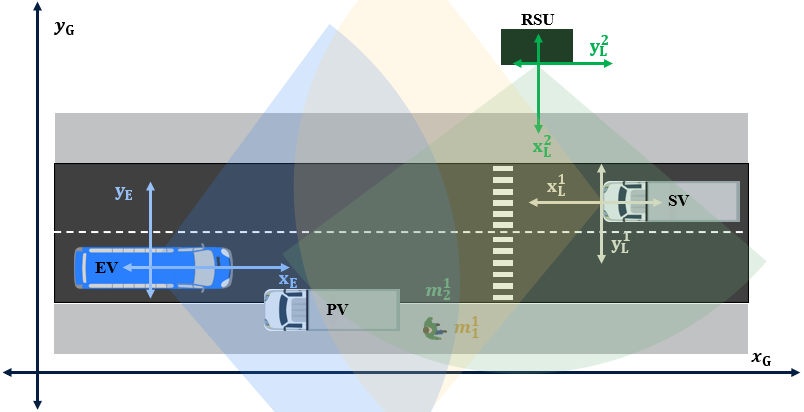}
    \caption{Scenario with four road-users and their coordinate frames.}
    \label{fig:coordinates_pf}
    \vspace{-4mm}
\end{figure}

\subsection{Scenario Description}

The animated version of the considered scenario is presented in Fig.~\ref{fig:coordinates_pf}. This scenario consists of five road-users, which are an EV, a Side Vehicle (SV), a Parked Vehicle (PV), a pedestrian, and an RSU. The FOVs for EV, SV, and RSU are presented in blue, yellow, and green circle segments, respectively. The PV is located in the scenario in such a way that it creates an occlusion in the FOV of the EV. The pedestrian, also called the unprotected agent, is in the occluded region. In this case, the understanding of the surrounding environment is limited to the road elements that are observable within the FOV of the EV. Behavior learning and prediction in this setting are limited by the time interval for which such elements are locally observable. Hence, the risk assessment tasks for a short decision-making horizon can only be made through a proactive manner by lowering speed to be able to handle any road-user that might appear in the occluded area. 
The perception information is shared using Collaborative Perception Message (CPM) service. A detailed description of the CPM service is given in Section~\ref{subsec:data}. 

In Fig.~\ref{fig:coordinates_pf}, it can be seen that various coordinate frames are involved when it comes to real-time implementation. A global coordinate system is the earth-fixed coordinate frame, which is represented in meters in north and east directions from an agreed reference point, which is represented by $\text{x}_G$ and $\text{y}_G$, respectively. 
Each agent has a local coordinate frame with respect to the reference point. These reference points are their global positions in the global coordinate frame. The EV has its local coordinate frame with axes $\text{x}_E$ and $\text{y}_E$. The SV has its local coordinate frame with axes $\text{x}_{L}^1$ and $\text{y}_{L}^1$. Similarly, for RSU, $\text{x}_{L}^2$ and $\text{y}_{L}^2$ are its local coordinate frame axes. These different types of coordinates frames are presented in Fig.~\ref{fig:coordinates_pf}. The PV in the EV's FOV is detected as a static obstacle, creating an occlusion in EV's FOV as shown in Fig.~\ref{fig:coordinates_pf}. The pedestrian is in the occluded region of the EV's FOV. Therefore the EV can not detect the pedestrian. But whereas, SV and RSU can detect it in their FOVs.



\subsection{Proposed Architecture}
\begin{figure}[t]
    \centering
    \includegraphics[width=0.95\linewidth]{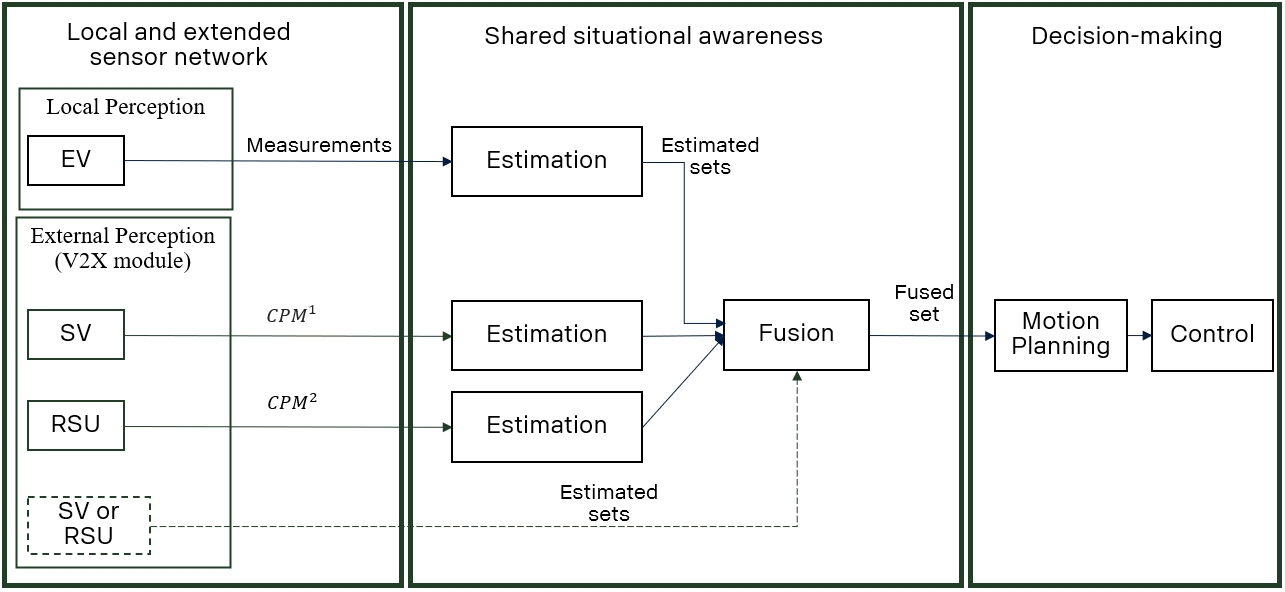}
    \caption{Shared situational awareness internal architecture.}
    \label{fig:arch_v2}
    \vspace{-4mm}
\end{figure}

The proposed architecture consists of (i) Local and extended sensor networks, (ii) Algorithms for shared situational awareness, and (iii) Decision-making, as presented in Fig.~\ref{fig:arch_v2}. In the local and extended sensor network part, the blocks EV, SV, and RSU represent the road-users capable of providing perception data with respect to their local coordinate frame. We fuse the data on two levels: (i) measurement data and (ii) estimated sets. This two-level data fusion method helps us build a modular system, which means that if there is another SV or RSU in the scenario capable of sharing estimated sets, then these estimated sets can be fused directly with other estimated sets in the fusion module. Based on these fused sets, decisions can be made, and actions can be planned. In this paper, the main focus is on the sensor network and shared situational awareness.

\subsection{Local and External Perception Data} \label{subsec:data}

\label{subsubsec:cpmservice}
\begin{figure}[t]
    \centering
    \includegraphics[width=1\linewidth]{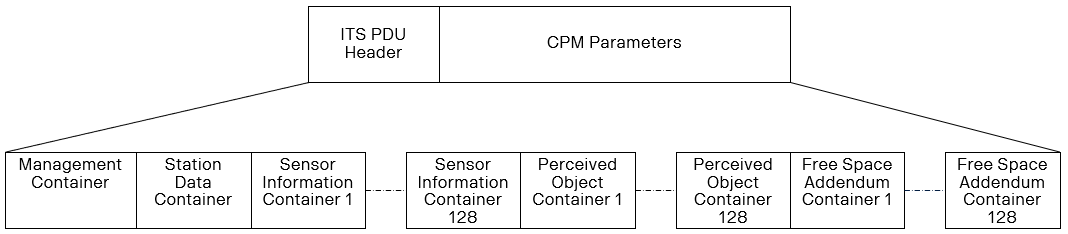}
    \caption{General structure of CPM.}
    \label{fig:cpm_structure}
   \vspace{-6mm}
\end{figure}
The external perception data is obtained by using the CPM service. This service is implemented by using standards published by The European Telecommunications Standards Institute (ETSI)~\cite{ETSI_TR_103_562}. The general message structure for a CPM is as specified in the standards, which is shown in Fig.~\ref{fig:cpm_structure}. In this work, we use Management Container (MC), Station Data Container (SDC), and Perceived Object Container (POC). The MC provides basic information about the originating ITS-S (Intelligent Transport System-Station), which includes the station type and reference position in the global coordinates. The SDC provides more specific information about the originating ITS-S in addition to the common information provided by the MC. The data received from the perception module of SV or RSU is populated into the POC of the CPM. The POC enables a detailed description of the dynamic state and properties of a detected object. 

The perception data from the SV or RSU in the scenario has different coordinate frames depending on their position and orientation with respect to global coordinates. For instance, when a road-user is detected in the FOV of SV, the perception data is denoted by $m_{i,k}^j$ as shown in Fig.~\ref{fig:coordinates_pf}, denoting that it is $j^{th}$ road-user detected by $i^{th}$ V2X unit. This data is obtain from the respective perception layer of the V2X unit and information obtain from $m_{i,k}^j \in \mathbb{R}^{1 \times b}$ with $b=6$ as shown below: 
\begin{equation}
    \label{eq:measurement_from_preception_layer}
 m^j_{i,k} = \begin{bmatrix}
     m_\text{x} & m_\text{y} & m_\text{s} & \theta & l & w
 \end{bmatrix}^j_{i,k},
\end{equation}
where $m_\text{x}$ and $m_\text{y}$ are the relative distance from $i^{th}$ V2X unit at a given time stamp indicated by $k$ and the location of $i^{th}$ V2X unit is given in terms of global coordinates. The $m_\text{s}$ and $\theta$ are the velocity and heading of the detected $j^{th}$ road-user at $k^{th}$ time step. The size of the of road-user is indicated by length and width, i.e., $l$ and $w$. This perception data, $m^j_{i,k}$ is converted to $y^j_k$ with respect to the EV coordinate system with the help of the global coordinate of originating V2X unit and the global coordination of the EV.

In this work, it is considered to have all the local and external perception data with respect to the local coordinate frame of the EV to maintain uniformity. The local perception also has the same perception data structure as explained in~\eqref{eq:measurement_from_preception_layer}. The local perception data is already given with respect to the EV's local coordinate frame, which means that it does not require conversion of the coordinate system and can be realized in the format of $y^j_k$. It is considered that the EV reference point is at the center of the vehicle, and the reference points of SV and RSU are at the front of the vehicles, which creates an offset between different perception measurements as shown in Fig.~\ref{fig:coordinates_pf}. This offset has to be handled systematically to be able to fuse local and external perceptions. 



\subsection{Algorithm for Shared Situational Awareness}

Our shared situational awareness algorithm consists of two steps: (1) Set-based estimation and (2) Fusion. These steps are explained below and summarized in Algorithms \ref{alg:prediction} and \ref{alg:fusion}.

\subsubsection{Set-Based Estimation}

The set-based estimation approach computes a set that contains the true state with guarantees. 
The prediction set is estimated with the aid of the system state model expressed in~\eqref{eq:sysmodel}. The process and measurement noise for $j^{th}$ road-user are assumed to be unknown but bounded by zonotopes: $q_k^j \in \mathcal{Z}_{{Q},k}^j = \langle0,Q_k\rangle$, and $v_k \in \mathcal{Z}^j_{R,k}=\zono{0,\text{diag}(r^j_k)}$. Then, we intersect the aforementioned predicted set with the set that aligns with the measurement set. In short, we have the following three sets and the aim is to compute the corrected state set $\bar{\mathcal{Z}}^j_{k} \subset \mathbb{R}^n$.

\begin{definition}[\textbf{Predicted State Set}]
\label{def:predset}
Given the system in~\eqref{eq:sysmodel} and \eqref{eq:senmodel}, the initial set $\mathcal{Z}^j_{0}= \langle c^j_{0},G^j_{0} \rangle$, and the process noise zonotope $\mathcal{Z}^j_{Q,k}$, the predicted reachable set of states $\hat{\mathcal{Z}}^j_{k} \subset \mathbb{R}^{n}$ is:
\begin{equation}
\hat{\mathcal{Z}}^j_{k}= F^j_k \bar{\mathcal{Z}}^j_{k-1} \oplus \mathcal{Z}^j_{Q,k}. 
\end{equation}
\end{definition}

\begin{definition}[\textbf{Measurement State Set}] 
\label{def:measset}
Given the system in~\eqref{eq:sysmodel} and \eqref{eq:senmodel}, and the measurement noise zonotope $\mathcal{Z}^j_{R,k}=\zono{0,\text{diag}(r^j_k)}$ where the measurement state set $\mathcal{S}^j_{k} \subset \mathbb{R}^p$ of the $j^\text{th}$ road-user, is the set of all possible solutions $x^j_{k}$ which can be reached given the measurement $y^j_{k}$. Note that when $y^j_{k} \in \mathbb{R}^{p}$ is a scalar, i.e., $p=1$, this measurement set is a strip:
\begin{equation}
    \mathcal{S}^j_{k} = \Big\{ x^j_{k} \Big| | H x^j_{k} - y^j_{k}| \leq r^j_k \Big\}. \label{eq:strip}
\end{equation}
\end{definition}
\begin{definition}[\textbf{Corrected State Set}]
\label{def:corrset}
Given the system~\eqref{eq:sysmodel} and \eqref{eq:senmodel} with initial set $\mathcal{Z}^j_{0}= \langle c^j_{0},G^j_{0} \rangle$, the reachable corrected state set $\bar{\mathcal{Z}}^j_{k}$ for the $j^\text{th}$ road-user is defined as:
\begin{equation}
     \big( \hat{\mathcal{Z}}^j_{k} \cap \mathcal{S}^j_{k} \big) \subseteq \bar{\mathcal{Z}}^j_{k}. 
\end{equation}
\end{definition}

Set-based approaches intersect the set of states consistent with the model  (predicted state set), denoted by $\hat{\mathcal{Z}}^j_{k-1}$, and the sets consistent with the measurements  (measurement state set), denoted by $\mathcal{S}^j_{k}$, $j = 1,\dots,n_m$, to obtain the corrected state set, denoted by $\bar{\mathcal{Z}}^j_{k}$, which is an over-approximation of the resultant intersection. This corrected set is also known as estimated set. The strips represent measurements from the sensors and are fused together to get a corrected state set. We use the following proposition to fuse the measurements from the sensors and extended sensors.

\begin{proposition}[\cite{conf:stripzono}]
 \label{th:stripzono}
 Given are zonotope $\hat{\mathcal{Z}}^j_{k-1}= \langle \hat{c}^j_{k-1},$ $\hat{G}^j_{k-1} \rangle $, the family of $n_m$ measurement sets $\mathcal{S}^j_{k}$ in \eqref{eq:strip} and the design parameters $\lambda^j_{k} \in \R^{n \times p}$, $j = 1,\dots,n_m$. The intersection between the zonotope and measurement sets can be over-approximated by the zonotope $\bar{\mathcal{Z}}^j_{k}=\langle  \bar{c}^j_{k},\bar{G}^j_{k}\rangle $, where
 \begin{align}
  \bar{c}^j_{k} &=  \hat{c}^j_{k-1} + \sum\limits_{j=1}^{n_m} \lambda^j_{k}(y^j_{k} - H^j \hat{c}^j_{k-1}), \label{eq:C_lambda}\\
  \bar{G}^j_{k} &= \Big[ (I - \sum\limits_{j}^{n_m} \lambda^j_{k} H^{j} ) \hat{G}_{k-1}, \lambda^1_{k} r^1_{k},\dots,\lambda^{n_m}_{k} r^{n_m}_{k} \Big]. 
  \label{eq:G_lambda}
 \end{align}
  \end{proposition}

The design parameter $\lambda^j_{k}$ can be obtained by solving an optimization problem to minimize the volume of the resultant zonotope~\cite{conf:dis-diff}. After fusing the measurements from the sensors, we fuse the estimated sets from multiple vehicles and infrastructure units. Set-membership estimation is performed on each measurement separately and the algorithm is as shown in~\ref{alg:prediction}.

\begin{algorithm}[ht]
\caption{Set-based estimation for each road-user}\label{alg:prediction}
\KwInput{Initial set ${\mathcal{Z}}_{0} = \langle c_0, G_0 \rangle$, Process noise zonotope $\mathcal{Z}_{Q,k} = \langle0,Q_k\rangle$, Measurement noise zonotope $\mathcal{Z}_{R,k} = \langle0,R_k\rangle$, Measurement $y^j_k$, 
%
}
\KwOutput{Corrected state set $\bar{\mathcal{Z}}^j_{k} = \langle c^j_{k}, G^j_{k} \rangle$, $\forall j =1,\dots,n_m$, $k=1,\dots,T$ } 

\For{ $j = 1,\dots,n_m$}{



 
Compute the predicted set $ \hat{\mathcal{Z}}^j_{k}= F^j_k \bar{\mathcal{Z}}^j_{k-1} \oplus \mathcal{Z}^j_{Q,k}$.

Collect measurement set: $\mathcal{S}^j_{k} =\{H^j, y^j_{k},r^j_k\}$.

Compute the corrected set: $ \bar{\mathcal{Z}}^j_{k} \supseteq \big( \hat{\mathcal{Z}}^j_{k} \cap \mathcal{S}^j_k \big) $. 

        
 }

\end{algorithm}

\subsubsection{Fusion}


 We propose to use a fusion function in order to fuse our internal estimated sets with the received estimated sets. More specifically, consider when a road-side unit or connected vehicle is reporting estimated sets of the same pedestrian. Then, we make use of the following proposition to fuse the estimated sets $\bar{\mathcal{Z}}^j_{k}$, $j = 1,\dots,n_e$, by finding their intersections. 
 \begin{proposition}[\cite{conf:dis-diff}]
 \label{th:diff}
The intersection between $n_e$ zonotopes $\bar{\mathcal{Z}}^j_{k}=\zono{\bar{c}^j_{k} , \bar{G}^j_{k}}$ can be over-approximated using the zonotope $\grave{\mathcal{Z}}^j_{k}=\zono{\grave{c}^j_{k},\grave{G}^j_{k}} \subset \mathbb{R}^n $ as follows:
 \begin{eqnarray}
\grave{c}^j_{k}&=&\frac{1}{\sum\limits_{j}^{n_e} w^j_{k}}\sum \limits_{j}^{n_e}w^j_{k}\bar{c}^j_{k},\label{eq:cdiff}\\
\grave{G}^j_{k} &=& \frac{1}{\sum\limits_{j=1}^{n_e} w^j_{k} }[w^1_{k}\bar{G}^1_{k} ,...,w^{n_e}_{k} \bar{G}^{n_e}_{k}], \label{eq:gdiff}
 \end{eqnarray}
 where $w^j_{k}$ is a weight such that $\sum\limits_{j}^{n_e} w^j_{k} \neq 0$.
 \end{proposition}

\begin{figure}[t!]
    \centering
    \includegraphics[width=0.6\linewidth]{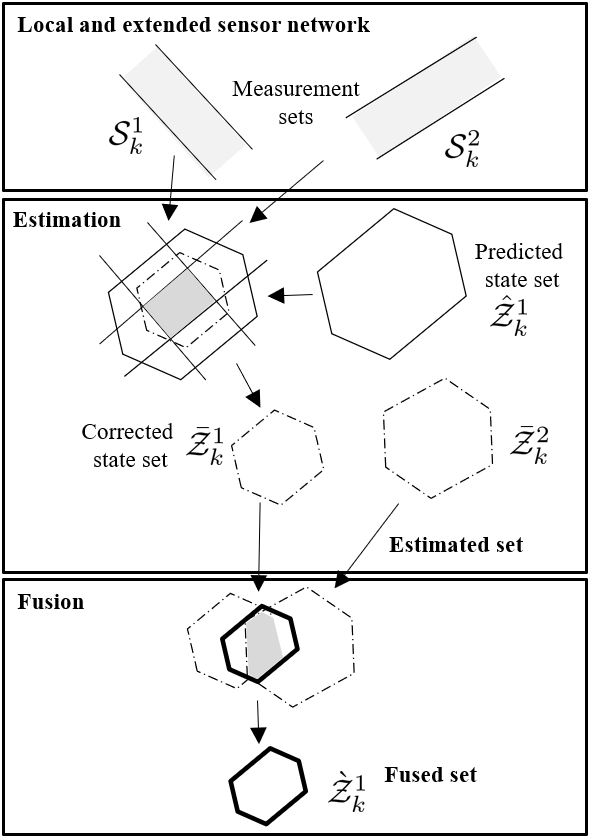}
    \caption{Set-based estimation and fusion.}
    \label{fig:zonostripinter}
    \vspace{-4mm}
\end{figure}

Again, the design parameter $w^j_{k}$ can be obtained by solving an optimization problem to minimize the size of the resultant zonotope \cite{conf:dis-diff}. The algorithm for set-based fusion is described in Algorithm~\ref{alg:fusion}. Estimation and fusion of the shared situational awareness algorithm are described with illustrations in Fig.~\ref{fig:zonostripinter}. In Fig.~\ref{fig:zonostripinter}, $\bar{\mathcal{Z}}^2_{k}$ can be estimated set from another V2X unit near the EV.

\begin{algorithm}
\caption{Shared set-based fusion for each user}\label{alg:fusion}

\KwInput{List of estimates : $\bar{\mathcal{Z}}^j_{k}$ for $j=1,\dots,n_e$}

\KwOutput{$\grave{\mathcal{Z}}^j_{k}=\zono{\grave{c}_{k},\grave{G}_{k}}$}

If a road-user has multiple estimates zonotopes i.e. $j = 1,2..,n_e$ and $\bar{\mathcal{Z}}^j_{k}=\zono{\bar{c}^j_{k} , \bar{G}^j_{k}}$, then compute: 

$\grave{c}_{k} = \frac{1}{\sum\limits_{j}^{n_e}w^j_{k}}\sum\limits_{j}^{n_e}w^j_{k}\bar{c}^j_{k}$;

$\grave{G}_{k} = \frac{1}{\sum\limits_{j=1}^{n_e}w^j_{k}}[w^1_{k}\bar{G}^1_{k} ,...,w^{n_e}_{k}\bar{G}^{n_e}_{k}]$;

 $\grave{\mathcal{Z}}_{k}=\zono{\grave{c}_{k},\grave{G}_{k}}$;
 
\label{eq:gdiff_alg}
\end{algorithm}

\section{Evaluation}\label{sec:evaluation}

In this section, we start by describing our experimental setup, then followed by presenting results. 

\subsection{Experimental Setup}\label{sec:ExSet-up}

\begin{figure}[t]
    \centering
    \includegraphics[width=\linewidth]{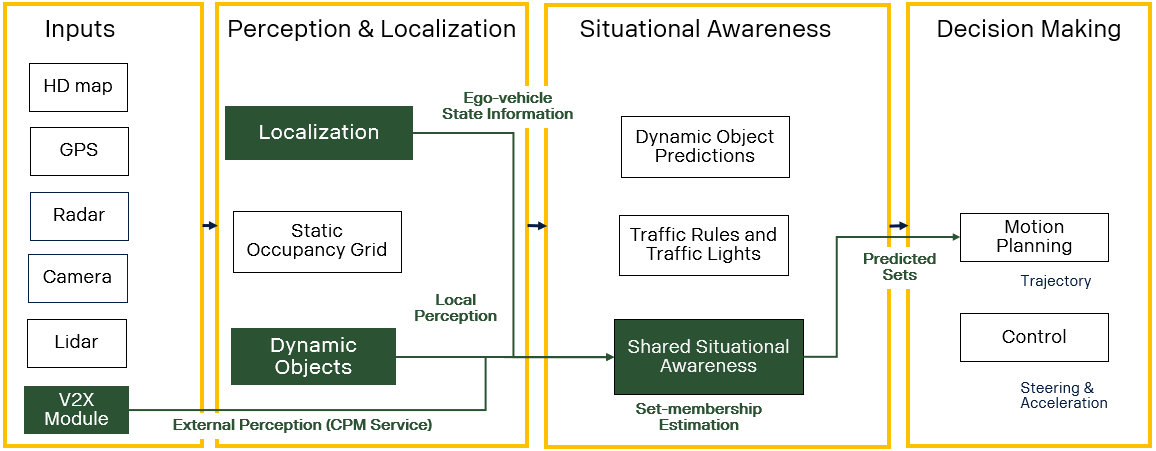}
    \caption{Shared Situational Awareness integrated to the existing system architecture.}
    \label{fig:ssaw_with_existingArch}
    \vspace{-4mm}
\end{figure}

In this section, details from the test experimental-up will be highlighted. The experiments were carried out at the Scania test track. For these tests, three CAVs and a parked vehicle were used. These three CAVs were equipped with various sensors and were capable of computing their own local perception. In Fig.~\ref{fig:ssaw_with_existingArch}, integration of the shared situational awareness algorithm with the existing system architecture of the EV is demonstrated. The EV state information and local perception data are extracted from the localization module and the dynamic object module. In order to receive CPM messages with external perception from other road-users, a V2X communication system was installed and configured. 

\begin{figure}[t]
    \centering
    \includegraphics[width=0.57\linewidth]{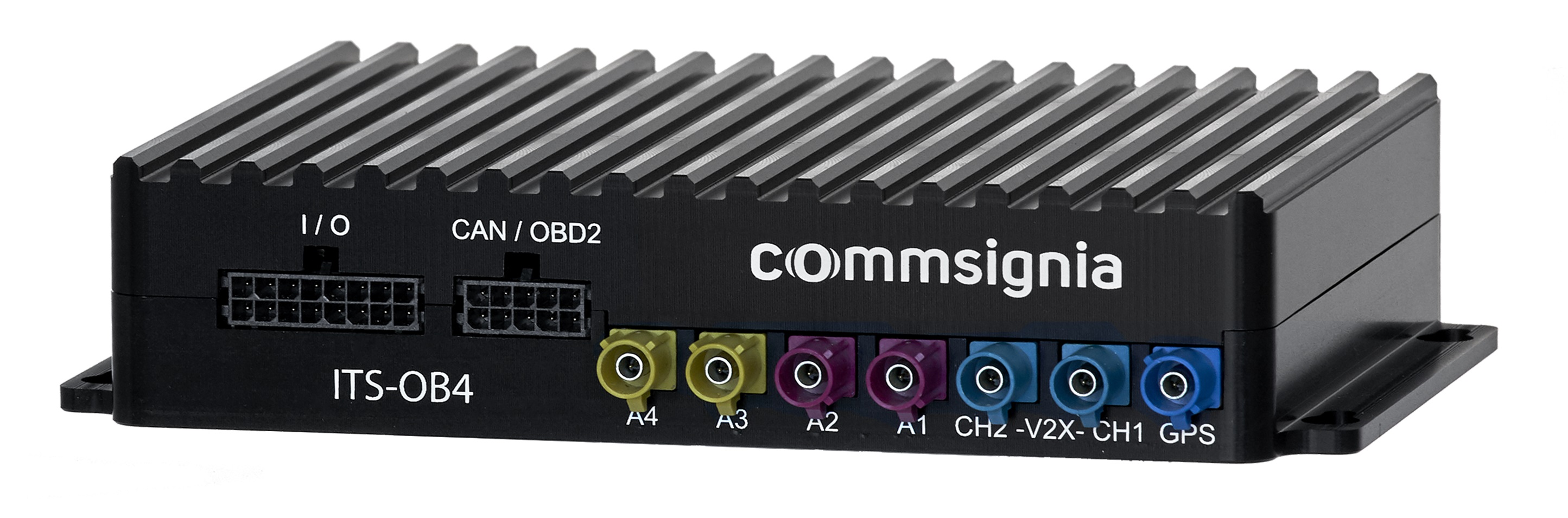}
    \caption{Commsignia ITS-OB4; vehicular connectivity system used in this project.}
    \label{fig:v2x_module}
    \vspace{-4mm}
\end{figure}

In this project, Commsignia ITS-OB4, the fourth generation of vehicular connectivity system, enables vehicular communication, presented in Fig.~\ref{fig:v2x_module}. This system communicates to other connected agents in the scenario using Dedicated Short-Range Communications (DSRC), which is based on ITS-G5. ITS-G5 is a European standard for vehicular communication based on IEEE 802.11 standards for its lower layers.

For the test performed in this project, the frequency used for communication is 5.900 GHz. The range varies a lot depending on the operating area. On an open field, the range is more than 1000 meters at the line of sight, but if the receiver is hiding behind something, it is possible the sender is not heard, even if they are close to each other. On the other hand, in an urban environment with many buildings, the signal will bounce on the buildings and therefore be available to receivers around corners and heard while there is no line of sight. This makes it powerful in crowded areas. The capacity is measured to receive around 500 msgs/sec; we transmit each message type at 10 Hz today.


\begin{figure*}[t]
    \centering
    \begin{subfigure}[b]{0.4\textwidth}
        \centering
        \includegraphics[width=1\linewidth,height=0.4\linewidth]{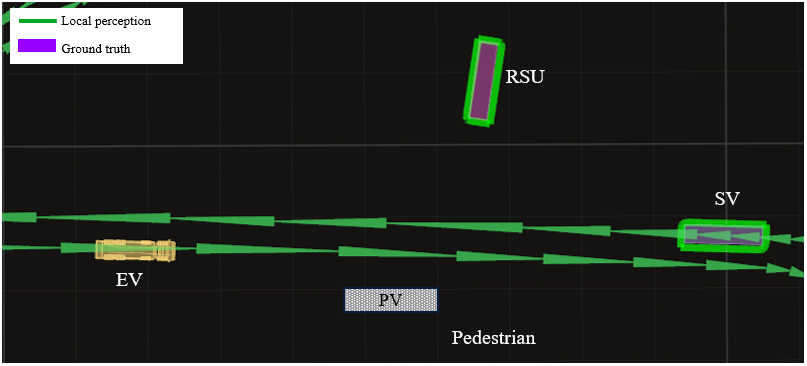}
        \caption{Scene map with ground truth of SV and RSU and estimated sets from local perception which are represented by green zonotopes.}
        \label{fig:results_lp}
     \end{subfigure}
     \begin{subfigure}[b]{0.4\textwidth}
        \centering
        \includegraphics[width=1\linewidth,height=0.4\linewidth]{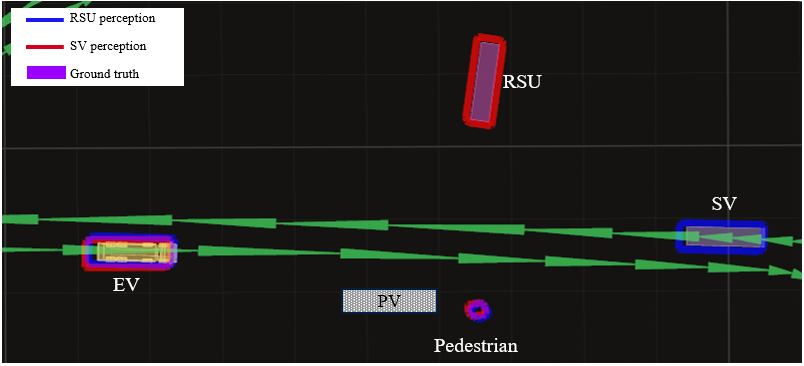}
        \caption{Scene map with estimated sets from external perception of SV and RSU which are represented by red and blue zonotopes respectively.}
        \label{fig:results_sv_and_rsu}
        
     \end{subfigure}
           \begin{subfigure}[b]{0.4\textwidth}
        \centering
        \includegraphics[width=1\linewidth,height=0.4\linewidth]{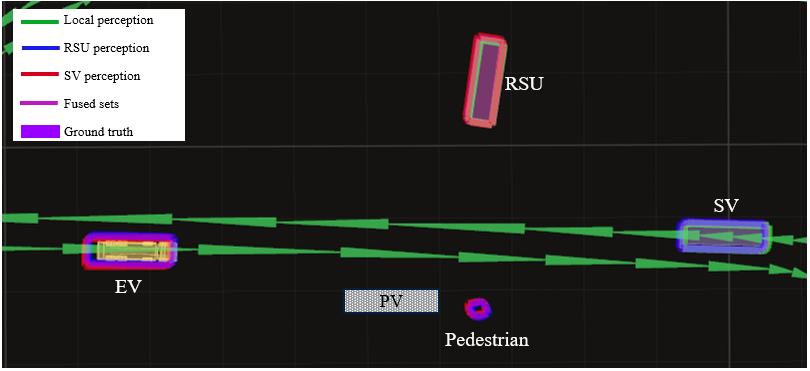}
        \caption{Scene map with estimated sets from local and external perception and fused sets.}
        \label{fig:results_total_scene_map}
     \end{subfigure}
     \begin{subfigure}[b]{0.4\textwidth}
        \centering
        \includegraphics[width=1\linewidth,height=0.4\linewidth]{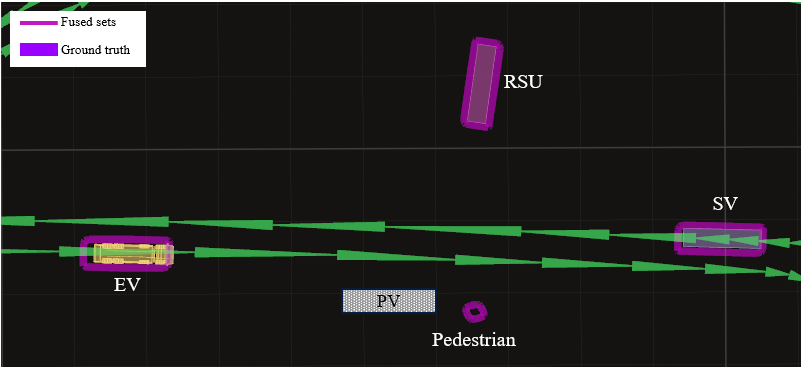}
        \caption{Scene map with fused sets which are represented by purple zonotopes.}
        \label{fig:results_diffusion}
     \end{subfigure}
     
     \caption{Real time results from the test run of the given scenario.}
     \label{fig:Simulation_results}
     \vspace{-4mm}
\end{figure*}

\begin{figure*}[ht!]
    \centering
    \begin{subfigure}[b]{0.4\textwidth}
        \centering
        \includegraphics[width=1\linewidth,height=0.5\linewidth]{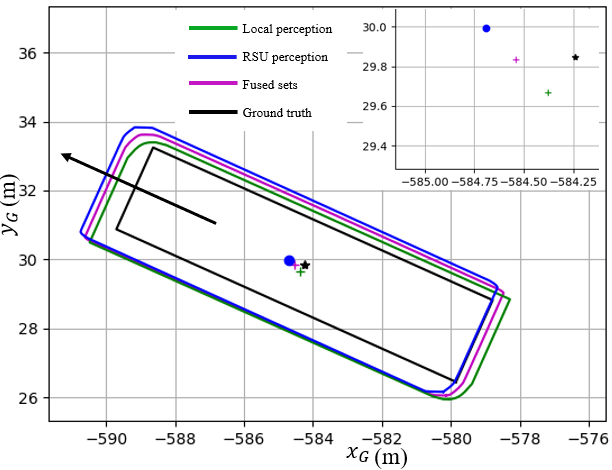}
        \caption{Estimated sets from local and external perception and the ground truth of SV, along with respective centers.}
        \label{fig:matlab_plot_sv}
     \end{subfigure}
     \begin{subfigure}[b]{0.4\textwidth}
        \centering
        \includegraphics[width=1\linewidth,height=0.5\linewidth]{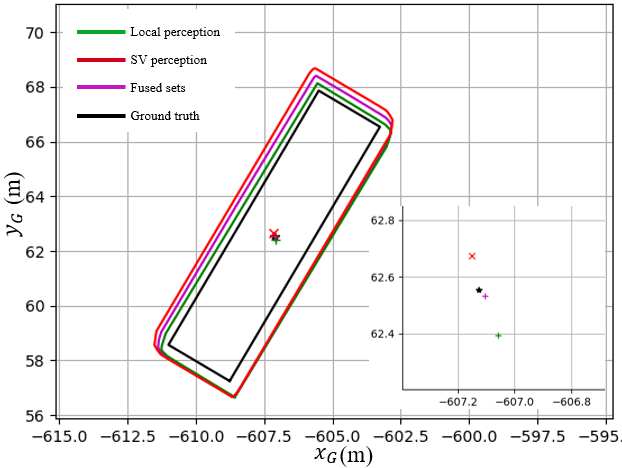}
        \caption{Estimated sets from local and external perception and the ground truth of RSU, along with respective centers.}
        \label{fig:matlab_plot_rsu}
     \end{subfigure}
      \begin{subfigure}[b]{0.4\textwidth}
        \centering
        \includegraphics[width=1\linewidth,height=0.5\linewidth]{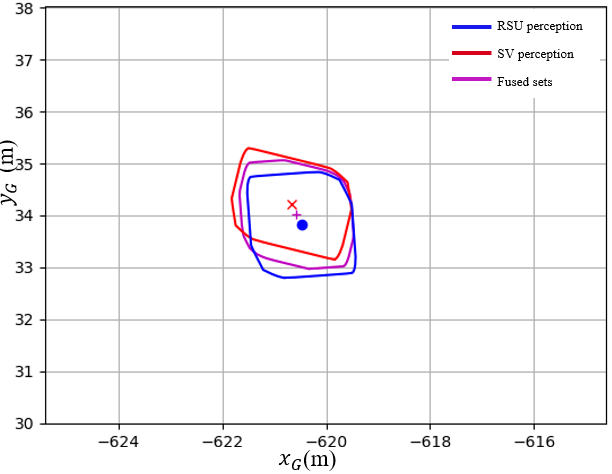}
        \caption{Estimated sets from local and external perception of EV, along with respective centers.}
        \label{fig:matlab_plot_ped}
     \end{subfigure}
     \begin{subfigure}[b]{0.4\textwidth}
        \centering
        \includegraphics[width=1\linewidth,height=0.5\linewidth]{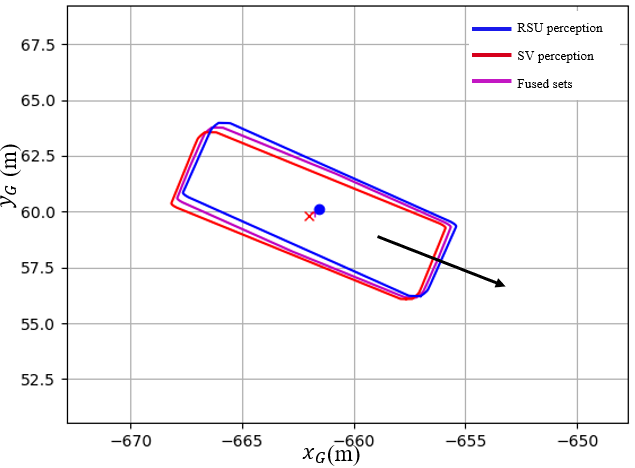}
        \caption{Estimated sets from local and external perception of pedestrian, along with respective centers.}
        \label{fig:matlab_plot_ev}
     \end{subfigure}
     \caption{Matlab plot of real time results of each actor.}
     \label{fig:matlab_plots_individual}
     \vspace{-4mm}
\end{figure*}

\begin{table*}[t]
    \centering
    \caption{Comparison of ground truth, estimated set and fused set of the SV and RSU in the given scenario.}
    \begin{tabular}{|c|c|c|c|c|c|c|c|c|}
    \hline
     \multirow{2}{*}{Actor} &
      \multicolumn{2}{c|}{Ground truth} &
      \multicolumn{2}{c|}{Estimated set from} &
      \multicolumn{2}{c|}{Estimated set from}&
       \multicolumn{2}{c|}{Fused set}\\
       \multirow{2}{*}{} &
      \multicolumn{2}{c|}{} &
      \multicolumn{2}{c|}{local perception} &
      \multicolumn{2}{c|}{external perception}&
       \multicolumn{2}{c|}{}\\
       & Center $(m,m)$ & Area $m^2$& Center $(m,m)$ & Area $m^2$ & Center $(m,m)$ & Area $m^2$ & Center $(m,m)$ & Area $m^2$\\
        \hline
        SV & $(-584.24,29.85)$& $28.08$&$(-584.38,29.67)$ &$40.80$& $(-584.70,29.99)$&$42.92$ &$(-584.54,29.84)$&$42.04$\\
        \hline
        RSU & $(-607.13,62.55)$& $28.08$ &$(-607.06,62.40)$ &$38.18$ & $(-607.15,62.67)$ &$43.64$ &$(-607.10,62.53)$&$41.13$\\
        \hline
    \end{tabular}
    \label{tab:estimation_ground_truth_comparison}
    \vspace{-4mm}
\end{table*}

\subsection{Results}\label{sec:results}

In this section, we will discuss the outcome of the test run of the scenario presented in Fig.~\ref{fig:scenario_realtime}. Fig.~\ref{fig:Simulation_results} presents various snapshots of EV's scene map. The EV's scene map consists of (i) the EV's location, which is represented by the yellow truck, (ii) a map with lane markings, (iii) the ground truth of SV and RSU, and (iii) estimated sets. The green arrows represent the lane center and direction, and the arrow's size is the lane's segmentation. It can also be observed that the PV is not included in the local perception, as PV is considered a static obstacle and is combined with the occupancy grid map, which contains information on infrastructure and occlusion in EV's FOV. 

In Fig.~\ref{fig:results_lp}, the green polygons represent the estimated set of the detected objects by the local perception of EV. In this scene map, it can be seen the PV occludes the pedestrian. 
Two agents in the scenario are capable of sharing perception data using V2X communication, i.e., SV and RSU. The global truth of these agents is sent via the CPM message, which can be further used for comparison. In Fig.~\ref{fig:results_sv_and_rsu}, the red polygon represents the estimated set from the SV perception, and the blue polygon represents the estimated set from RSU perception with reference to the EV's coordinate frame. Both SV and RSU are able to detect all the road-users in the given scenario from their FOV. 

After collecting all estimated sets from EV, SV, and RSU perception, these zonotopes are then grouped together if they have an intersection that indicates that they belong to the same road user and then fused using~\eqref{alg:fusion}. The resulting fused sets, along with local and external perception, are presented in Fig.~\ref{fig:results_total_scene_map}. In Fig.~\ref{fig:results_diffusion}, we present the fused set along with the ground truth of SV and RSU. This information about the road-users in the occluded regions, along with the set size, can be used in the motion planning module of EV. The improvement in the scene map can be observed when we compare Fig.~\ref{fig:results_lp} and Fig.~\ref{fig:results_diffusion}.

To be able to analyze the sets, the results have to be plotted using the MATLAB plotting tool in Fig.~\ref{fig:matlab_plots_individual}. With the help of another V2X communication service, i.e., the Cooperative Awareness Message (CAM) service, we obtain the ground truth of both SV and RSU. In Fig.~\ref{fig:matlab_plot_sv}, the estimated sets from local perception and RSU perception are presented with green and blue zonotopes, respectively, along with the fused set using purple zonotope. These estimations are compared with the ground truth of the SV, which is presented with a black rectangle. 
On the top right corner of Fig.~\ref{fig:matlab_plot_sv}, the centers of all estimated and fused sets, along with the ground truth, are presented. Similarly, in Fig.~\ref{fig:matlab_plot_rsu}, the estimated set from local perception and SV are presented with green and red zonotopes, respectively and along with the fused set presented using purple zonotope. These estimations are compared with the ground truth of the RSU, which is presented by the black rectangle. The centers of the estimated and the fused sets are compared with the ground truth, which is presented in the lower right corner of Fig.~\ref{fig:matlab_plot_rsu}.

In this scenario Fig.~\ref{fig:scenario_realtime}, the pedestrian is in the FOV's SV and RSU without an obstruction; therefore, they are detected by SV and RSU. In Fig.~\ref{fig:matlab_plot_ped}, the estimated set from the external perception from SV and RSU are presented by red and blue zonotopes, respectively, and the purple zonotope represents the fused set. The intersecting area of both blue and red zonotopes ideally will have the pedestrian, given the uncertainties in the measurements, which is further over-approximated to the convex zonotopes presented by the purple zonotope. Similarly, estimated and fused sets of the EV can be observed in Fig.~\ref{fig:matlab_plot_ev}.

In table~\ref{tab:estimation_ground_truth_comparison}, the numerical values of various zonotopes' centers of SV and RSU are compared with their respective ground truths. The area of each zonotope is also calculated and compared with each other. In the case of SV, the RMSE (Root Mean Square Error) between ground truth and estimated sets from local perception and external perception are $0.228$ and $0.480$, respectively. Similarly, in the case of RSU, the RMSE between ground truth and estimated sets from local perception and external perception are $0.165$ and $0.122$, respectively. The RMSE between ground truth and fused set for SV and RSU are $0.300$ and $0.036$, respectively.

\section{Conclusions}\label{sec:conc}

The proposed framework was successfully implemented on CAV's existing system architecture. In the given scenario with occlusion, we assumed that there are two V2X units, i.e., a stationary and a dynamic V2X unit, close to the crossing pedestrian. These V2X units share external perceptions with EV, and this data is processed by using the proposed framework to obtain safety guarantees. Real-time perception data from both EV and V2X units were used to perform test runs. During the real-time tests, the level of uncertainty in the measurements was realized. The proposed framework can fuse multiple measurements from both local and external perceptions. Improvement in the scene map of EV can be observed by comparing Fig.~\ref{fig:results_lp} and Fig.~\ref{fig:results_diffusion}. The set-membership method provides state estimation along with guarantees. 
These guarantees are essential for safety-critical applications. An immediate next step is to find a way to quantitatively present the improvement of EV's situational awareness while using the shared situational algorithm. 



\section*{Acknowledgment}
This work was partially supported by the Wallenberg Artificial Intelligence, Autonomous Systems, and Software Program (WASP) funded by the Knut and Alice Wallenberg Foundation. It was also partially supported by the Swedish Research Council, Swedish Research Council Distinguished Professor Grant 2017-01078 Knut and Alice Wallenberg Foundation Wallenberg Scholar Grant and Swedish Strategic Research Foundation CLAS Grant RIT17-0046. We would like to thank Henrik Pettersson from Scania CV AB for his valuable insights and support.

\bibliographystyle{ieeetr}
\bibliography{ref.bib}

\begin{thebibliography}{10}

\bibitem{wang2019multi}
Z.~Wang, Y.~Wu, and Q.~Niu, ``Multi-sensor fusion in automated driving: A
  survey,'' {\em Ieee Access}, vol.~8, pp.~2847--2868, 2019.

\bibitem{cui2022cooperative}
G.~Cui, W.~Zhang, Y.~Xiao, L.~Yao, and Z.~Fang, ``Cooperative perception
  technology of autonomous driving in the internet of vehicles environment: A
  review,'' {\em Sensors}, vol.~22, no.~15, p.~5535, 2022.

\bibitem{yoon2021performance}
D.~D. Yoon, B.~Ayalew, and G.~M.~N. Ali, ``Performance of decentralized
  cooperative perception in v2v connected traffic,'' {\em IEEE Transactions on
  Intelligent Transportation Systems}, 2021.

\bibitem{InformationEntropy}
Z.~Chen, L.~Xiong, and C.~Tang, ``Perception-aware path planning for autonomous
  vehicles in uncertain environment,'' tech. rep., SAE Technical Paper, 2022.

\bibitem{mo2022method}
Y.~Mo, P.~Zhang, Z.~Chen, and B.~Ran, ``A method of vehicle-infrastructure
  cooperative perception based vehicle state information fusion using improved
  kalman filter,'' {\em Multimedia Tools and Applications}, vol.~81, no.~4,
  pp.~4603--4620, 2022.

\bibitem{queralta2019collaborative}
J.~P. Queralta, T.~N. Gia, H.~Tenhunen, and T.~Westerlund, ``Collaborative
  mapping with ioe-based heterogeneous vehicles for enhanced situational
  awareness,'' in {\em 2019 IEEE Sensors Applications Symposium (SAS)},
  pp.~1--6, IEEE, 2019.

\bibitem{CombastelSurfaceVehicles}
L.~Orihuela, C.~Combastel, and G.~Bejarano, ``Set-membership state estimation
  of autonomous surface vehicles with a partially decoupled extended
  observer,'' in {\em 2022 European Control Conference (ECC)}, pp.~2024--2029,
  2022.

\bibitem{yee2018collaborative}
R.~Yee, E.~Chan, B.~Cheng, and G.~Bansal, ``Collaborative perception for
  automated vehicles leveraging vehicle-to-vehicle communications,'' in {\em
  2018 IEEE Intelligent Vehicles Symposium (IV)}, pp.~1099--1106, IEEE, 2018.

\bibitem{9709015}
R.~Yu, D.~Yang, and H.~Zhang, ``Edge-assisted collaborative perception in
  autonomous driving: A reflection on communication design,'' in {\em 2021
  IEEE/ACM Symposium on Edge Computing (SEC)}, pp.~371--375, 2021.

\bibitem{9575828}
V.~Narri, A.~Alanwar, J.~Mårtensson, C.~Norén, L.~Dal~Col, and K.~H.
  Johansson, ``Set-membership estimation in shared situational awareness for
  automated vehicles in occluded scenarios,'' in {\em 2021 IEEE Intelligent
  Vehicles Symposium (IV)}, pp.~385--392, 2021.

\bibitem{bolting2019iterated}
J.~Bolting and S.~Fergani, ``The iterated extended set membership filter
  applied to relative localization between autonomous vehicles based on gnss
  and uwb ranging,'' {\em Asian Journal of Control}, vol.~21, no.~4,
  pp.~1556--1565, 2019.

\bibitem{blesa2012robust}
J.~Blesa, V.~Puig, and J.~Saludes, ``Robust fault detection using
  polytope-based set-membership consistency test,'' {\em IET Control Theory \&
  Applications}, vol.~6, no.~12, pp.~1767--1777, 2012.

\bibitem{conf:dis-diff}
A.~Alanwar, J.~J. Rath, H.~Said, and M.~Althoff, ``Distributed set-based
  observers using diffusion strategy,'' {\em arXiv preprint arXiv:2003.10347},
  2020.

\bibitem{belforte1990parameter}
G.~Belforte, B.~Bona, and V.~Cerone, ``Parameter estimation algorithms for a
  set-membership description of uncertainty,'' {\em Automatica}, vol.~26,
  no.~5, pp.~887--898, 1990.

\bibitem{xu2020interval}
F.~Xu, S.~Yang, and B.~Liang, ``Interval set-membership estimation for
  continuous linear systems,'' {\em International Journal of Robust and
  Nonlinear Control}, vol.~30, no.~14, pp.~5305--5321, 2020.

\bibitem{conf:zono1998}
W.~K{\"u}hn, ``Rigorously computed orbits of dynamical systems without the
  wrapping effect,'' {\em Computing}, vol.~61, no.~1, pp.~47--67, 1998.

\bibitem{gonzalez_review_2016}
D.~Gonzalez, J.~Perez, V.~Milanes, and F.~Nashashibi, ``A review of motion
  planning techniques for automated vehicles,'' {\em {IEEE} Transactions on
  Intelligent Transportation Systems}, vol.~17, no.~4, pp.~1135--1145, 2016.

\bibitem{ETSI_TR_103_562}
ETSI, {\em Intelligent Transport Systems (ITS); Vehicular Communications; Basic
  Set of Applications; Analysis of the Collective Perception Service (CPS);
  Release 2}.
\newblock The European Telecommunications Standards Institute, {ETSI TR 103 562
  V2.1.1 (2019-12)}~ed., 2019.

\bibitem{conf:stripzono}
V.~T.~H. Le, C.~Stoica, T.~Alamo, E.~F. Camacho, and D.~Dumur, ``Zonotope-based
  set-membership estimation for multi-output uncertain systems,'' in {\em IEEE
  International Symposium on Intelligent Control}, pp.~212--217, 2013.

\end{thebibliography}
\end{document}